# What do we Really Know about State of the Art NER?

**Sowmya Vajjala, Ramya Balasubramaniam**
National Research Council, Canada; Novisto.com
sowmya.vajjala@nrc-cnrc.gc.ca, ramya.balasubramaniam@novisto.com

**Abstract**

Named Entity Recognition (NER) is a well researched NLP task and is widely used in real world NLP scenarios. NER research typically focuses on the creation of new ways of training NER, with relatively less emphasis on resources and evaluation. Further, state of the art (SOTA) NER models, trained on standard datasets, typically report only a single performance measure (F-score) and we don't really know how well they do for different entity types and genres of text, or how robust are they to new, unseen entities. In this paper, we perform a broad evaluation of NER using a popular dataset, that takes into consideration various text genres and sources constituting the dataset at hand. Additionally, we generate six new adversarial test sets through small perturbations in the original test set, replacing select entities while retaining the context. We also train and test our models on randomly generated train/dev/test splits followed by an experiment where the models are trained on a select set of genres but tested genres not seen in training. These comprehensive evaluation strategies were performed using three SOTA NER models. Based on our results, we recommend some useful reporting practices for NER researchers, that could help in providing a better understanding of a SOTA model's performance in future.

## 1. Introduction

NER is one of the most commonly researched information extraction tasks in NLP. It is also among the most common use-cases in industry, according to a recent survey (Lorica and Nathan, 2021). State of the art (SOTA) NER models report F-scores of over 90% on standard English datasets[1]. There are also several off-the-shelf NER tools that provide pre-trained models based on this research, which can be used in practical application scenarios.

However, what do we really know about these NER models, beyond a single measure (typically, micro-F score)? One approach to answer this question could be to study how well does an NER model do on various named entity categories and text genres, or how well does it generalize on data with previously unseen genres and new test sets. Another approach could be to assess how robust the algorithms themselves are when trained and tested on different splits of the same dataset. There is some recent research in NLP that shows why reporting and comparing based on a single evaluation measure is not the best practice (Reimers and Gurevych, 2018; Dror et al., 2019). However, there isn't much research focusing on NER in particular. Consequently, to understand the performance of SOTA NER models in depth, and argue for better evaluation approaches in future, we address the following research questions in this paper:

1. How well do pre-trained SOTA NER models perform
   - on various named entity categories?
   - on various data sources and text genres?

2. How robust are the pre-trained NER models if we replace the entities in test set with new, unseen entities of the same type?

3. How sensitive are the NER models to:
   - training on random splits of train/dev/test sets?
   - text genres not seen during training?

We address these questions not by proposing new NER models or new evaluation metrics, but by studying three well-known NLP libraries based on SOTA NER models, using a large, and commonly used multi-genre and multi-source NER dataset for English and conducting multiple rounds of evaluations. Based on our results, we list a few recommendations on how to comprehensively evaluate and report performance of SOTA NER models.

Considering that NER models are used as-is in many industry use cases, and that NLP research does not evaluate such aspects as a norm, we believe this paper provides useful insights for both NER researchers and practitioners, and hope this encourages more research in developing a holistic evaluation framework for NER systems in future.

The main contributions of this paper can be summarized as follows:

1. We perform an extensive evaluation of three SOTA NER approaches and identify key insights for developing better NER models.

2. We create six new adversarial test sets for English NER, by applying small perturbations on the original test set.

The rest of this paper is organized as follows: Section 2 gives an overview of contemporary work in this area and sets the stage for the work carried out in this paper.

---

[1] http://nlpprogress.com/english/named_entity_recognition.html

Section 3 describes our experimental methodology and dataset creation. Section 4 discusses our experiments that do "black-box" testing of three SOTA NER models. Section 5 describes results of our experiments with training and testing NER models. Section 6 highlights the main conclusions of the paper and recommendations for evaluating and reporting NER model performance.

## 2. Related Work

Research about Named Entity Recognition in NLP community has primarily focused on developing new models (typically for English), but to a much lesser extent on resource creation and evaluation. OntoNotes 5.0 is a large, multi-genre dataset that is commonly used for the development and evaluation of NER models. In OntoNotes benchmarking paper, Pradhan et al. (2013) hopes that this dataset "provides an opportunity for studying the genre effect on different syntactic, semantic and discourse analyzers". However, this is not really reflected in practice, at least in NER. SOTA NER models built using this dataset report just a single F score for English [2] (e.g., Yu et al. (2020) - 91.3; Bhattacharjee et al. (2020) - 92.07; Xu et al. (2021) - 90.85; Shah et al. (2021)-92.17). However, none of these papers shed light on how good the models really are, beyond this single F score.

Apart from reporting the overall F score, some papers discuss genre-wise performance (Pradhan et al., 2013; Chiu and Nichols, 2016; Ghaddar and Langlais, 2018; Bernier-Colborne and Langlais, 2020; Fu et al., 2021a; Fu et al., 2021b), and some go beyond that to explore cross-corpus/cross-genre evaluation (Wang et al., 2020; Bernier-Colborne and Langlais, 2020; Ushio and Camacho-Collados, 2021). However, we still do not have a clear picture of how existing architectures perform on genres unseen during training. Considering the practical relevance of NER, this is important aspect of performance that needs to be better understood.

A majority of NER research focuses on training and evaluating with standard train/dev/test splits, and there are problems associated with it such as overfitting to the test set. Although there have been calls in the recent past for evaluating on random splits (Gorman and Bedrick, 2019), using multiple test sets (Søgaard et al., 2021), and using a "tune-set" (van der Goot, 2021), we haven't seen this being reflected in NER research much. Though several papers report on experiments that build and evaluate NER models on more than one dataset (Ushio and Camacho-Collados, 2021; Bernier-Colborne and Langlais, 2020), they still primarily report results on the standard splits for a given dataset. Some papers do report variance across multiple runs of training (Strubell et al., 2017), but we haven't found much analysis of how the performance changes with non-standard splits or with (slightly modified) test sets for NER.

[2] https://paperswithcode.com/sota/named-entity-recognition-ner-on-ontonotes-v5

Developing generalizable NLP models is an open research question. NER research too focused on this aspect in the past, from a modeling perspective. For example, Wang et al. (2020) proposed methods develop models to account for multi-genre scenarios. Ghaddar et al. (2021) recently proposed an approach that forces models to learn more contextual information and memorize less. Zeng et al. (2020) generate counterfactual examples to enhance the original dataset while training a model. However, these approaches address only one side of the problem.

The other side of generalizability is about evaluating on new test sets. Performing adversarial attacks has been a recent area of research in NLP and machine learning to understand some aspects of this question. The goal of such an approach is to create minimal perturbations in the input text, which will maximize the probability of a model making wrong predictions. From simple word/character substitutions and heuristics targeting various forms of text transformations to using large pre-trained language models, a range of strategies have been proposed in the past, to generate adversarial text (Eger et al., 2019; Ren et al., 2019; Jin et al., 2020; Yangming et al., 2020; Keller et al., 2021).

In the recent past, such methods have been applied for various tasks such as co-reference resolution (Chai et al., 2020), question answering (Ravichander et al., 2021), morphological analysis (Shmidman et al., 2020), natural language inference (Wallace et al., 2019), named entity linking (Goel et al., 2021). Agarwal et al. (2020) extended this research to NER by creating new test sets replacing some entities selectively, using lists of common names by various nationalities. Lin et al. (2021)'s RockNER used wikidata and BERT to substitute entities and contexts and generate new NER test sets. In this paper, we extend this line of enquiry further by creating six new adversarial test sets for evaluating NER (in English).

To summarize, inspired by some past research into generalizability (Augenstein et al., 2017; Taille et al., 2020) and evaluation (Lignos and Kamyab, 2020; Bernier-Colborne and Langlais, 2020; Fu et al., 2020; Tu and Lignos, 2021; Xu et al., 2021) of NER, and evaluation of NLP systems in general (Ribeiro et al., 2020), we attempted to understand what we could learn about SOTA NER beyond a single F-score in this paper. While some of these questions were addressed separately in the past research to various degrees, to our knowledge, there is no previous research that addressed all these questions together, comparing multiple NER systems in the past.

## 3. Our Approach

In this paper, we conduct two kinds of experiments:

1. black box evaluation, where we evaluate pre-trained NER models for their performance on various entity types, data sources/genres, and new test sets.

2. training NER models using existing model architectures to understand their sensitivity to random splits and cross-genre performance.

Our research questions require a dataset with multiple genres of NER annotated text. Further, since our focus is more on understanding the current SOTA, we required some easily re-trainable implementations of SOTA NER models. Our methodological choices, which are described in this section, are motivated by these requirements.

### 3.1. Dataset

To our knowledge, Ontonotes 5.0 is the only NER dataset for English with multi source/genre annotation. Hence, we used it for the experiments described in this paper. The NER part of Ontonotes 5.0 is a large dataset with 3,637 documents and 2 million tokens (Pradhan et al., 2013). The NER annotation consists of 18 tags, including 11 types (Person, geo-political entity, organization etc) and 7 values (date, percent etc). The dataset contains texts from six sources: broadcast conversation (bc), broadcast news (bn), magazines (mz), newswire (nw), telephone conversation (tc), weblog (wb). We re-grouped these into four genres - news (bn + nw + mz), bc, tc, and wb and report on results both by source and genre. The distinction between news and broadcast conversation is just that while the former is primarily in written form, the latter is a written version of broadcast news conversations. We decided to group bn, mz and nw to a common category news as we did not see much linguistic variation amongst these categories and hence there was no value in treating them as separate genres.

### 3.2. New adversarial test sets

We created **six** new test sets by replacing entities from an entity type in the standard test, using the Python package Faker[3]. Our motivation for this experiment is to understand how much of memorization of entities occurs in NER models and how sensitive the SOTA NER models are to small changes or perturbations, i.e., replacing some entities to unseen values, while keeping the surrounding context intact. Faker package generates fake named entity data, based on geographical, gender and entity type specifications. We generated the following 6 test sets with input perturbations, and all except the first one were generated using Faker:

1. Perturb_1: Replace all person names with the word *Dodo*. The motivation for this simple perturbation is to check how much do the models learn to infer by using context compared to actual memorization of lexical tokens.

2. Perturb_2: Replace all person names using en_US locale in Faker.

3. Perturb_3: Replace all person names using en_IN locale (India) in Faker.

4. Perturb_4: Replace all person names with female names, using en_TH locale (Thailand) in Faker.

5. Perturb_5: Replace all person names with female names, using en_IN locale in Faker.

6. Perturb_6: Replace all GPE names with GPE names from en_IE (Ireland) locale.

Note that the goal of the perturbations is to introduce new entities not seen during training into the test set. Consequently, each Perturb_* dataset addresses one transformation. Hence, only some sentences of the original test set, which meet the individual perturbation's criteria, are altered in each new test set. For names with more than one word, we called First name function once and last name function for the rest. For GPE with more than one word, we called place suffix function for all words after the first word. The locale/gender combinations are arbitrary and Faker can be used with other locale/gender combinations as well. Our goal was just to explore a small sample from a large population of possibilities, to understand SOTA NER's sensitivity to small changes in input. We chose gender and geography as a starting point in this exploration. Whether this strategy can be reliably used to evaluate NER models in terms of racial/sexist bias is an interesting aspect for future exploration.

Table 1 shows how these perturbations were implemented on sentences from test set. While the locale and gender appropriateness of these functions is not reported in Faker documentation, the sentences we generated by replacing the entity tokens are all grammatically correct and hence they all serve as valid test sets for our purpose.

### 3.3. NER systems

We experimented with three SOTA NER systems, using their pre-trained models as well as by training our own models with their architectures. They are described below:

1. Spacy[4] uses a neural network based state prediction model (transition based parser) to do structured prediction for NER. It comes with several pipelines for English. We used the *en_core_web_trf* pipeline, which fine-tunes a RoBERTa-base (Liu et al., 2019) for this task. Its pre-trained model also gives maximum performance for NER on OntoNotes dataset compared to other models that Spacy provides, as reported on their website[5].

2. Stanza[6]'s NER architecture uses BiLSTM lay-

---

[3] https://faker.readthedocs.io/en/master/

[4] https://spacy.io/
[5] https://github.com/explosion/spacy-models/releases/tag/en_core_web_trf-3.2.0
[6] https://stanfordnlp.github.io/stanza/

| Perturb | Sentence |
|---|---|
| None (Original) | Faced with massive demonstrations and Russia's backing of **Kostunica**, he agreed to step down in October. |
| Perturb_1 | Faced with massive demonstrations and Russia's backing of **Dodo**, he agreed to step down in October. |
| Perturb_2 | Faced with massive demonstrations and Russia's backing of **Kevin**, he agreed to step down in October. |
| Perturb_3 | Faced with massive demonstrations and Russia's backing of **Arnav**, he agreed to step down in October. |
| Perturb_4 | Faced with massive demonstrations and Russia's backing of **Pol**, he agreed to step down in October. |
| Perturb_5 | Faced with massive demonstrations and Russia's backing of **Samaira**, he agreed to step down in October. |
| None (Original) | Previously we had a statistic, especially for the ringroads in Beijing. |
| Perturb_6 | Previously we had a statistic, especially for the ringroads in **Galway**. |

Table 1: Examples of sentences in perturbed test sets

ers with character and word-level representations followed by a CRF decoder (Qi et al., 2018). We used Stanza's NER model trained on OntoNotes for evaluating their approach, and used the combined pre-trained embeddings provided with Stanza's standard installation both for testing its pre-trained NER system as well as for training our own NER models.

3. SparkNLP[7] uses a CharCNNs-BiLSTM-CRF model for training an NER model. We used its *ontonotes-bert-base-cased*[8] model for our evaluation. SparkNLP's training routine did not support usage of custom validation sets and instead used a percentage of training set for validation (we chose 10%).

It is technically possible to use other existing NER libraries and/or the more recently published NER models. However, these three are the most used NLP libraries (Lorica and Nathan, 2021) that support off the shelf NER. They provide both SOTA pre-trained models as well as training routines to train our own NER models. Hence, we chose these three libraries, which report SOTA or near SOTA results for NER on OntoNotes dataset. We provide all our code for replication, which can be applied with any English NER model/dataset that follows the standard BIO dataset format.[9]

### 3.4. Experimental settings

We report our results in the following three experimental setups:

1. The pre-trained NER models and the genre-specific models we trained were evaluated with the standard OntoNotes 5.0 test split described in Pradhan et al. (2013)[10], separating it by source/genre where needed.

2. The pre-trained NER models were also evaluated using the six new test sets we created.

3. We trained and tested models using the ten randomly generated train/dev/test splits.

In the original OntoNotes splits, a set of documents were kept aside as test set and the NER test sentences were taken from those documents. Our random train/dev/test splits were generated at the sentence level on the entire corpus, keeping the proportions of the original train/dev/test splits.

**Evaluation** As is common in NER research, we reported micro-averaged F scores for entity-level NER. We used Seqeval (Nakayama, 2018), a Python compatible version of the standard conlleval script for sequence evaluation. Table 2 shows a summary of model performance as reported on respective library websites, and the score we obtained when we ran those models with standard OntoNotes test splits.

| Library | Reported | Obtained | Delta |
|---|---|---|---|
| Stanza | 88.8 | 88.71 | 0.01 |
| SpaCy | 90.0 | 89.09 | 0.91 |
| SparkNLP | 89.97 | 88.6 | 1.37 |

Table 2: Performance of OntoNotes NER models in the three NLP libraries

While the scores we obtained are very close in the case of Stanza, we notice a performance difference of around 1% between the scores reported on the website and our evaluation for the other two libraries. It is possible that the actual deployed model on these libraries was different from the model with best results, due to operational reasons. It is also possible that the evaluation scripts used by the libraries are slightly different from each other and from seqeval. However, it has to be noted that the goal of this paper is to understand the performance of these approaches at a more general level, considering various aspects, consistently, rather than replicating the exact results reported in the model descriptions.

---

[7] https://nlp.johnsnowlabs.com/
[8] https://nlp.johnsnowlabs.com/2020/12/05/onto_bert_base_cased_en.html
[9] https://github.com/nishkalavallabhi/SOTANER/
[10] Available at: http://cemantix.org/data/ontonotes.html

## 4. Experiments: Black-box Evaluation

Our black-box NER experiments tackled the first two research questions on how SOTA NER models perform on various named entity types, for various data sources/text genres, and how robust are they to small perturbations on standard test set.

### 4.1. NER performance by types

OntoNotes dataset consists of 18 entity types in total, with a very uneven distribution (in terms of mention counts) across the train/dev/test splits. Table 3 shows the F-scores for four most frequent and least frequent entity types in the dataset, using the three libraries we evaluated[11].

| Category | Stanza | Spacy | SparkNLP |
|---|---|---|---|
| Most-frequent entity types | | | |
| DATE | 86.55 | 85.51 | 85.54 |
| GPE | 95.2 | 95.36 | 95.61 |
| ORG | 87.44 | 90.476 | 87.53 |
| PER | 93.29 | 93.51 | 93.11 |
| Least-frequent entity types | | | |
| LANGUAGE | 60.61 | 74.42 | 60.60 |
| LAW | 64.79 | 67.50 | 64.71 |
| EVENT | 64.96 | 74.42 | 53.22 |
| PRODUCT | 67.97 | 71.95 | 71.05 |

Table 3: F-scores for 4-most and 4-least frequent entity types

As we can observe from Table 3, there is a huge performance difference among various entity types, and less frequent entity types show poorer results for all the three libraries. All three libraries show similar performance for more frequent entity types, but for the less frequent entity types they show noticeable differences (close to 20% for EVENT category for Spacy versus SparkNLP). A possible reason for this could just be that we need more training examples for the less frequent entity types, or it could be that these entity types are indeed difficult to tag or even difficult to annotate for humans. Yet, OntoNotes is perhaps the largest NER dataset available for English, and it could be challenging to further increase dataset size for increasing the representation of these entity types. Clearly, this issue needs a closer look as reporting a single F score per model does not give a full picture of how good SOTA NER is. There is evidently some road to travel in terms of getting to holistically higher performance across all entity types.

---
[11]Full list for all 18 types can be seen in the supplementary material

### 4.2. NER performance by data source and genre

To analyze the performance further, we split standard test set into subsets based on their source and genre as described in Section 3. Tables 4 and 5 show the performance of the pre-trained NER models separated by source and genre respectively.

| Source | Stanza | Spacy | SparkNLP |
|---|---|---|---|
| bn | 91.82 | 91.64 | 90.93 |
| mz | 85.97 | 88.72 | 87.73 |
| nw | 90.87 | 86.14 | 90.96 |
| bc | 88.35 | 91.55 | 87.59 |
| tc | 76.68 | 71.16 | 78.38 |
| wb | 81.2 | 82.81 | 80.11 |

Table 4: NER Performance by data source

| Genre | Stanza | Spacy | SparkNLP |
|---|---|---|---|
| News (bn, mz, nw) | 90.41 | 90.79 | 90.47 |
| bc | 88.35 | 88.72 | 87.59 |
| tc | 76.68 | 71.16 | 78.37 |
| wb | 81.2 | 82.81 | 80.11 |

Table 5: NER Performance by genre

We notice some performance difference across sources for all models (Table 4). The drop in F-score for best and worst performance across sources is ∼15% for Stanza, ∼20% for Spacy and ∼12% for SparkNLP. If we aggregate by genre (Table 5), the performance is higher for news, with a slight drop for bc, larger drops for wb and tc. The drop in F-score between the best and worst genre is ∼14% for Stanza, ∼20% for Spacy, and ∼12% for SparkNLP. This leads us to conclude that the SOTA NER approaches, while reaching F scores around 90%, are still not as good for genres such as web blogs and telephonic conversations, compared to the more commonly seen news genre.

### 4.3. Adversarial Test Sets

As described in Section 3, we created six new test sets from standard test splits by carefully replacing some entities with newly generated values, without changing the context. Table 6 shows performance of all three NER models on the six adversarial test sets, in comparison with the original test set (indicated by *None* setting).

For all person name perturbations (Perturb_1–5), overall NER performance did not drop drastically, with the largest drop being closer to 3% for all three models. The largest drop in all three models was for the test set with Perturb_4. While this drop is large enough to change the SOTA model, it need not be considered drastic in general. However, drop was much larger for PER category itself. It was over 10% for this category, for Perturb_4, across all three models. Stanza, Spacy and SparkNLP had F-scores of 93.29, 93.51, 93.11 for PER category respectively, with the original test set. However, they dropped to 80.75, 83.64, 80.61 respectively with Perturb_4 test set.

| Setting | Stanza-All | Stanza-PER | Spacy-All | Spacy-PER | SparkNLP-All | SparkNLP-PER |
|---|---|---|---|---|---|---|
| None | 88.71 | 93.29 | 89.09 | 93.51 | 88.6 | 93.11 |
| Perturb_1 | 88.66 | 93 | 88.29 | 91.51 | 86.21 | 82.98 |
| Perturb_2 | 87.8 | 88.18 | 88.75 | 92.13 | 88.14 | 90.48 |
| Perturb_3 | 87.95 | 90.35 | 88.34 | 90.84 | 88 | 91.22 |
| Perturb_4 | 85.57 | 80.75 | 86.74 | 83.64 | 85.24 | 80.61 |
| Perturb_5 | 87.88 | 90.15 | 88.42 | 91.37 | 87.84 | 90.26 |
|  | Stanza-All | Stanza-GPE | Spacy-All | SpacyGPE | SparkNLP-All | SparkNLP-GPE |
| None | 88.71 | 95.2 | 89.09 | 95.61 | 88.6 | 95.61 |
| Perturb_6 | 80.87 | 68.37 | 82.48 | 73.47 | 80.56 | 68.84 |

Table 6: Performance of the NER systems with adversarial test sets

Spacy seemed relatively more robust to all the other PER category perturbations with <3% drop across the other four perturbations. Stanza's model resulted in a 5% drop for Perturb_2, but otherwise, within 3% for others. Strangely enough, SparkNLP showed a drop of >10% for Perturb_1, where all person names were replaced with the word *Dodo*. Clearly, the models seem to be memorizing some of the entities that appear both in training and test set. When we look only at overall F-score, it is hard to identify NER models' sensitivity to such seemingly minor changes to input.

In the case of Perturb_6, performance drop was more noticeable, both for overall F as well as F for GPE category. While we would need further analysis to understand the reason for a more significant drop as compared to other perturbations, a possible explanation could be that there was larger token-level overlap between original test set and training data for GPE entities (75.7%) compared to PER entities (61.2%). Further, percentage of unique tokens was lower for GPE (∼9%) than PER (∼20%), even within training set. Hence, it is possible that models are memorizing the tokens than learning to infer using context, and this could be occurring more for GPE entity type than for other entity types. Thus, GPE could be more vulnerable than PER category, when tested with adversarial inputs.

Overall, these experiments with black-box testing of NER models showed that SOTA NER models we explored did not perform equally well across different NE categories, performed inconsistently across data sources/genres, and are sensitive to seemingly small input perturbations, which only introduced new entity values without changing context. We believe that this sensitivity to small perturbations particularly warrants further investigation into what exactly do NER models learn. Further, it has to be noted that we explored a small set of possible perturbations, and only for 2 of the 18 entity/numeric tags in the dataset. Future research could focus on other tags and other input perturbations, to get a fuller picture on what the NER system actually learns apart from memorization.

## 5. Experiments: Training NER

We performed two experiments involving training of NER models. The first one trained NER models using 10 randomly generated train/dev/test splits, and the second experiment explored a trained NER model's performance on unseen genres. In the latter, we preselected the genres used in training, development and testing steps within the OntoNotes dataset.

### 5.1. Random Splits

As described in Section 3, we generated 10 random train/dev/test splits, keeping their proportions same as in the standard split. Table 7 summarizes NER model performance, in terms of average, minimum and maximum micro F-score.

| Model | Avg. F | S.Dev | min | max |
|---|---|---|---|---|
| Stanza | 88.34 | 0.37 | 87.77 | 89.09 |
| Spacy | 90.77 | 1.17 | 87.98 | 92.19 |
| SparkNLP | 90.04 | 0.20 | 89.6 | 90.41 |

Table 7: Performance of NER with random splits

There is only a 1-1.5% variation in the performance across splits for Stanza and SparkNLP, but a 4.2% variation for Spacy. Considering that the chosen split can actually result in such difference in performance among the three chosen libraries, we conclude that in absence of multiple test sets, it would be good to evaluate an NER approach by training multiple models with random splits, to understand whether a model's performance is specific to the composition of the splits.

We performed a two-tailed paired sample t-test[12] over the F-scores for the 10 splits to understand whether there is a significant difference in performance across the three models. While the performance of Spacy and SparkNLP are not significantly different from each other, but were both significantly better than Stanza (p < 0.01).

---

[12] https://www.socscistatistics.com/tests/ttestdependent/default2.aspx

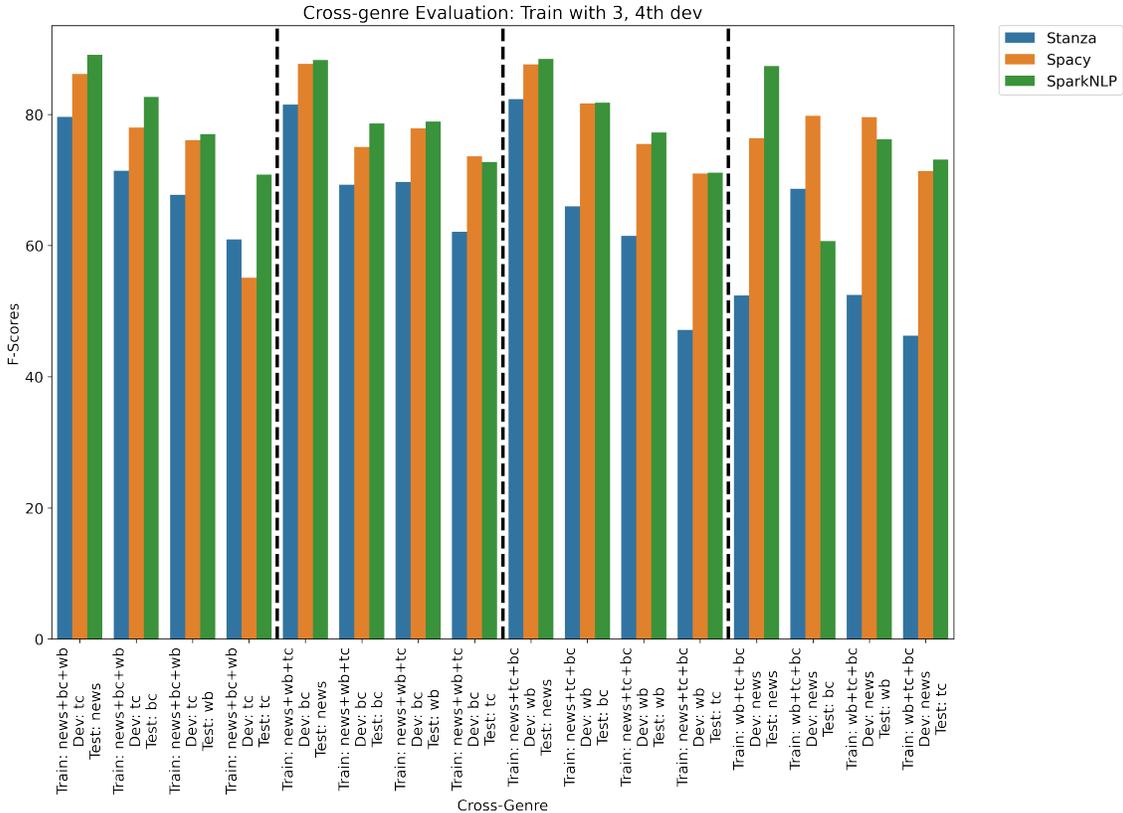

Figure 1: Training on Multiple-genres

## 5.2. Cross-Genre performance

To evaluate whether SOTA NER approaches can work well in a cross-domain setting, we trained NER models, in two settings:

1. Training on a single genre, and testing on all four.
2. Training on any three genres, and testing on all four.

We also explored genre-specific parameter tuning by changing development set each time.

**Training on a single genre** Since "news" was the largest subset of all, we used that for training and evaluated these models on test sets from all four genres (which is nothing but the original OntoNotes test set split into four subsets). Table 8 summarizes the results of this experiment, by training on news-train and tuning on news-dev[13].

From Table 8, we observed that all NER models performed poorly when used with new genres unseen during training. This performance drop was over 30% in the worst case (Spacy on tc). Even for the best performing approach, the drop was close to 10% when we evaluated on a new genre (SparkNLP on bc) which was the closest to the training genre (news) compared to others. Choosing genre specific development set either did not result in a major change in performance or resulted in slight drop in performance for that genre. Performance on genre-specific subsets for thesse genre-specific models was much lower than that for models trained on entire data (Table 5). This led us to conclude that SOTA training approaches won't give SOTA performance on unseen genres, even if we chose the best hyper-parameters for that genre based on development set.

| Model | news | bc | tc | wb |
|---|---|---|---|---|
| Stanza | 89.18 | 78.55 | 67.19 | 76.04 |
| Spacy | 82.64 | 65.4 | 51.64 | 62.57 |
| SparkNLP | 89.47 | 78.78 | 63.08 | 75.62 |

Table 8: Trained on Single genre

**Training on multiple genres:** Since our dataset had 4 genres, we explored whether performance on unseen genres would be better if the training data consisted of multiple genres. For this, we trained NER models by combining training sets of any three genres, and used development set of fourth genre. We then tested each of these models on all four genres. Our results are summarized in Figure 1.

---
[13]The results when we chose other genre development sets are provided in the supplementary material.

In all three libraries, we observed that performance across genres was lower than when NER models were trained using all four genres (Table 5). Although there was a consistent performance drop across all genres, news genre performed the best (right most in Figure 1). tc genre seemed to perform the worst (left most in the figure). When we compared this with single genre training, performance seemed slightly better, perhaps because training data were richer in each case (three genres versus one). Overall, it seemed that NER models did not perform well on genres unseen during training.

To summarize this set of experiments with training NER, we can conclude that evaluation across random splits indicate variation in performance, and the NER performance on genres unseen during training seems to a challenge. At this point, "quantity" of data seems to dominate "diversity", at least with this dataset.

## 6. Summary and Discussion

In this paper, we conducted several experiments to understand performance of three SOTA English NER approaches, using the OntoNotes dataset. Our findings can be summarized as follows:

1. The three SOTA models performed very differently across various NE categories, with huge performance variation observed among the entity types in the dataset.

2. The performance was also inconsistent across genres. For all models, performance on telephone conversation(tc) and web blog (wb) genres was much lower than that on the rest.

3. All models we explored were very sensitive to small input perturbations. The performance difference was starker in specific cases (Perturb_4 and Perturb_6).

4. Re-training and evaluating the three existing NER models on 10 randomly generated splits instead of the standard splits showed that there was sometimes over 4% variation in performance across the splits.

5. Single and multi-genre training and evaluation showed that SOTA NER models performed poorly on genres unseen during training, even when training data had multiple genres.

While we chose only three pre-trained English models and explored a limit set of evaluation dimensions, we believe that our observations are more general, and not model or architecture or language specific. Based on aforementioned findings, we recommend the following reporting practices for NER research:

1. Provide detailed performance table for the best performing model in the appendix, and report min/max individual F-scores along with averaged score for all reported models in the main content.

2. Perform experiments on other systematically/randomly generated splits apart from standard train/dev/test split and report any observed deviations.

3. Report results on multiple test sets to understand whether NER models just memorize entities or learn to tag using the context information.

4. Report results by source/genre too along with averaged performance, to understand the sensitivity of NER models to particular subsets of the dataset.

### 6.1. Outlook

Our experiments revealed several interesting aspects of NER system performance, which need further exploration. Identifying and addressing the effects of various forms of biases in NER datasets such as the ones introduced by source, genre, gender, region etc is an important aspect, following our experiments with adversarial test sets. Exploring adversarial test sets further, including other entity tags in the dataset would be useful to understand NER system robustness more comprehensively.

Improving the performance on under-represented NER categories and unseen text genres is another interesting problem to explore further. Considering the fact that our experiments have been conducted on a single language (English) and dataset(Onto Notes 5.0), we believe that we have a long way to go in terms of developing a more comprehensive, dataset and language agnostic evaluation approach.

Other related issues we did not touch include identifying annotation errors/inconsistencies in datasets (Reiss et al., 2020), error analysis of NER models (Stanislawek et al., 2019), development of new evaluation metrics (Bernier-Colborne and Langlais, 2020) and interpretable evaluation (Fu et al., 2020). We hope that the results and resources generated through this paper would lead to further research and development of a comprehensive evaluation suite for SOTA NER.

### 6.2. New Language resources

We created new test sets based on the original OntoNotes test set and also generated 10 new splits for train/dev/test files. OntoNotes is licensed under the terms of LDC[14] and hence, we cannot host the corpus publicly ourselves. However, we will share these resources with the research groups who have a license to use OntoNotes 5.0.

**Supplementary material:** https://github.com/nishkalavallabhi/SOTANER/.

---

[14] https://catalog.ldc.upenn.edu/LDC2013T19

# Acknowledgements

We would like to thank all the four reviewers, Taraka Rama, Gabriel Bernier-Colborne and Yunli Wang for their detailed feedback.

## 7. Bibliographical References